\documentclass[runningheads]{llncs}

\usepackage{caption}
\usepackage{graphicx}
\usepackage{amsmath}
\usepackage{amssymb}
\usepackage{booktabs}
\usepackage{enumitem}
\usepackage{mathrsfs}
\usepackage{booktabs}
\usepackage{multirow}
\usepackage{subcaption}
\usepackage[symbol]{footmisc}

\newcommand{\eg}[1]{\textit{i.e.}}

\usepackage[pagebackref,breaklinks,colorlinks,hyperfootnotes=false]{hyperref}

\usepackage[capitalize]{cleveref}
\crefname{section}{Sec.}{Secs.}
\Crefname{section}{Section}{Sections}
\Crefname{table}{Table}{Tables}
\crefname{table}{Tab.}{Tabs.}

\begin{document}

\title{Neural Capture of Animatable 3D Human from Monocular Video}

\pagestyle{headings}
\mainmatter


%
\author{
Gusi Te\inst{1,2}\thanks{Work done during an internship at Microsoft Research Asia.} \and
Xiu Li\inst{2,3}$^*$ \and
Xiao Li\inst{2} \and
Jinglu Wang\inst{2} \and
Wei Hu\inst{1}\thanks{Corresponding authors.} \and
Yan Lu\inst{2}$^\dagger$
}
\authorrunning{Tegusi et al.}
\institute{Peking University \and
Microsoft Research Asia \and
Tencent}
\maketitle

\begin{abstract}
    We present a novel paradigm of building an animatable 3D human representation from a monocular video input, such that it can be rendered in any unseen poses and views.
    Our method is based on a dynamic Neural Radiance Field (NeRF) rigged by a mesh-based parametric 3D human model serving as a geometry proxy.
    Previous methods usually rely on multi-view videos or accurate 3D geometry information as additional inputs; besides, most methods suffer from degraded quality when generalized to unseen poses.
    We identify that the key to generalization is a good input embedding for querying dynamic NeRF: A good input embedding should define an injective mapping in the full volumetric space, guided by surface mesh deformation under pose variation.
    Based on this observation, we propose to embed the input query with its relationship to local surface regions spanned by a set of geodesic nearest neighbors on mesh vertices. By including both position and relative distance information, our embedding defines a distance-preserved deformation mapping and generalizes well to unseen poses.
    To reduce the dependency on additional inputs, we first initialize per-frame 3D meshes using off-the-shelf tools and then propose a pipeline to jointly optimize NeRF and refine the initial mesh.
    Extensive experiments show our method can synthesize plausible human rendering results under unseen poses and views. 
\end{abstract}


\section{Introduction}
\label{sec:intro}
The problem of digital reconstruction, modeling and photo-realistic synthesis of humans from a video sequence such that it can be rendered with any pose from any viewpoint is important, which enables various applications ranging from character animation for games and movies to immersive experience for virtual conferencing.
This problem is extremely challenging due to the complicated joint space of human geometry, appearance, and dynamic motion given only RGB videos as observation, especially for monocular videos where multi-view concurrency is unavailable.

\begin{figure}[t]
    \centering
    \includegraphics[width=0.7\textwidth]{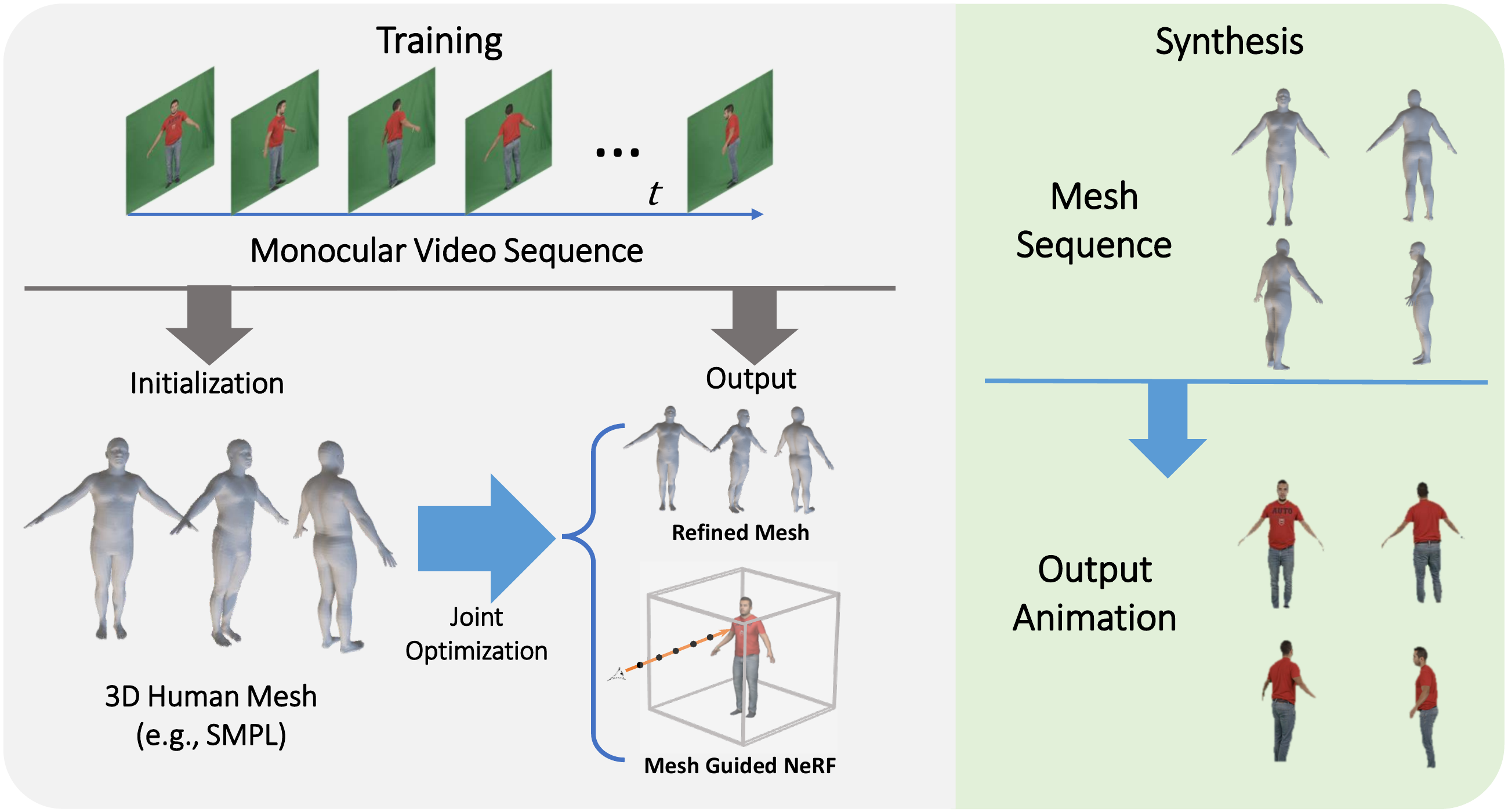}
    \caption{Left: given a monocular video sequence of human performance with initial posed 3D human with off-the-shelf tools, our method jointly reconstructs a mesh-guided neural radiance field (NeRF) and refined per-frame human mesh. Right: our trained mesh-guided NeRF is rigged with 3D mesh model and enables novel pose and view synthesis.}
    \label{fig:overview}
\end{figure}

Because of the difficulty in jointly modeling shape, pose and appearance of 3D humans from monocular videos, many previous approaches focus on solving part of the problem only, such as skeleton-based human pose estimation~\cite{he2020epipolar,cao2019openpose} or parametric 3D model~\cite{SMPL:2015,bogo2016keep} based human shape reconstruction~\cite{kolotouros2019learning,xiang2019monocular}.
These methods exploit sophisticated pose and shape priors and are thus able to partially counteract the geometry ambiguity; however, due to the lack of appearance information, the obtained results might not perfectly align with the input observations in certain frames. Extracted texture based on the estimated surface is usually blurry and cannot be used for photo-realistic synthesis.

Recently proposed volumetric neural rendering methods, \eg, NeRF and its variants~\cite{mildenhall2020nerf,sitzmann2020implicit,barron2021mip}, have shown great advances in high-quality free-view synthesis for static objects. NeRF models static objects by an implicit radiance field function with multi-layer perceptron (MLP) networks. Inspired by NeRF, recent works~\cite{peng2021neural,liu2021neural,peng2021animatable,noguchi2021neural} attempt to model 3D humans by conditioning the radiance field on 3D poses / parametric meshes.
While promising human reconstruction and view synthesis results have been achieved, these methods only focus on the modeling of conditional radiance field itself and require accurate 3D poses or meshes as a prior. This assumption is often too strong to be fulfilled in practical capture setups, especially with monocular video only.

To this end, we propose a novel paradigm of modeling an animatable 3D human representation from a monocular video sequence of a single person.
Our goal is to build a reconstruction pipeline with few non-trivial requirements such as accurate 3D human poses and/or geometry.
To achieve this goal, we propose to jointly optimize per-frame human mesh reconstruction and a dynamic neural radiance field (NeRF) which is conditional on mesh information.
Given a monocular video sequence as input observations, the optimization process is driven by the re-rendering error on the neural rendering output corresponding to both the NeRF and human poses, which are updated via back-propagation.
To constrain the optimization space of human mesh, we exploit the widely-used parametric human body model~\cite{bogo2016keep}, and initialize the optimization with poses provided by monocular pose estimation solutions~\cite{xiang2019monocular,kolotouros2019learning} as a starting point.
Our joint optimization strategy connects the (previously mangled) 3D geometry estimation problem and NeRF-based appearance optimization problem, and eliminates the requirement of accurate 3D geometry information as a priori,
making the modeling pipeline more applicable under monocular video scenarios.

A key property of a good neural representation of humans is that it should have good generalization under unseen human poses after training on limited observations.
This is a non-trivial task as previous NeRF-based works for human modeling \cite{peng2021neural,peng2021animatable} suffer from degraded quality more or less when generalized to unseen human poses.
Our observation is that the key for better pose generalization lies in the embedding method of input for querying NeRF.
Intrinsically, the dynamic NeRF-based representation of humans can be regarded as a static NeRF under rest pose equipped with 3D volume deformation that is conditioned on the mesh deformation from rest pose to any arbitrary target pose.
Thus, a good embedding for querying a dynamic NeRF input under arbitrary poses should "reverse" the pose deformation in an injective way to find the correct point at the static NeRF.
As the "correct" deformation mapping is only available on the surface mesh, the reverse deformation at any off-surface region in the space should be constrained with additional priors. 
Otherwise, the deformation mapping will be distorted and collapsed, thus failing to generalize to unseen poses.
\begin{figure*}[t]
    \centering
    \includegraphics[width=0.9\textwidth]{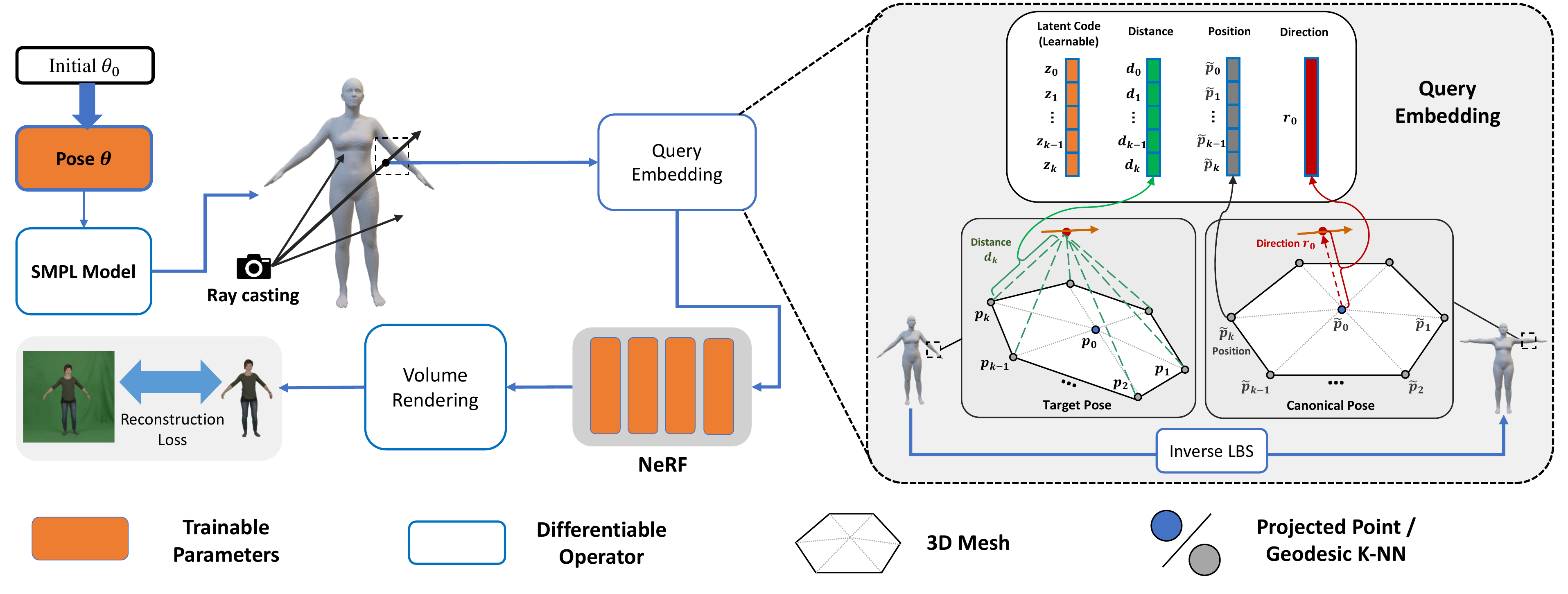}
    \caption{Our pipeline. A 3D mesh is generated from SMPL model with target pose $\theta$, followed with a mesh-guided NeRF which takes query embedding of 3D points and renders image via volume rendering. The query embedding encodes both surface deformation constraint with information of nearest mesh vertices under rest pose, as well as distance-preserve prior with distance to mesh vertices in a local region under target pose. During training the pose is initialized with off-the-shelf tools and are jointly refined with mesh-guided NeRF.}
    \label{fig:pipeline}
\end{figure*}

Based on this observation, we propose a new embedding method for querying mesh-guided dynamic NeRF by encoding the input position with its relationship to local nearby surface regions. Specifically, given a query point and a human mesh corresponding to a target pose, we project the query point onto the mesh and find a set of nearest neighbor mesh vertices locally; we then construct the input embedding with distances to these vertices in the \textbf{target} space as well as the normalized position of these vertices in the \textbf{canonical} space with \textbf{rest} human pose, eliminating pose deformation and view transformation.

Out proposed embedding method is able to guide the volume deformation at off-surface points with nearby surface deformation (as we give the inverse-transformed nearest neighbor vertices on mesh). It has two key properties that are essential for improving generalization.
First, the embedding is locally based on a nearby small connected region on the guided mesh. The local priors are crucial because they prevent the network from inadvertently relating the output to irrelevant articulated parts, which is known to hurt model generalization to poses unseen during training ~\cite{zeng2020srnet,noguchi2021neural}.
Second, since we give the distances to all nearest neighboring vertices in the target space, the embedding will encourage a locally distance-preserve prior to restrain the deformation from collapse.

Our method requires only the monocular video of a single person with a fixed camera, which does not rely on dedicated capture devices and/or accurate human pose information.
Extensive experimental results demonstrate the superiority of our model on a variety of data that exhibit various human shapes and poses.
To summarize, our contributions are as follows:
\begin{itemize}
    \item We propose a novel paradigm for building a neural human representation that can be rendered in unseen poses and views with monocular video inputs.
    \item We propose a novel input embedding representation for querying mesh-guided NeRF which improves the generalization ability on novel poses.
    \item We develop a pipeline for joint optimization of 3D human meshes and mesh-guided dynamic NeRF supervised by the reconstruction loss only.
\end{itemize}

\section{Related Works}
\label{sec:related_works}
\noindent{\textbf{Human Reconstruction.}}
The problem of digital reconstruction of humans is a long-standing problem in computer vision and computer graphics.
Traditional methods usually achieve high quality with complicated capture setups such as multi-view capture studio~\cite{dou2016fusion4d,zheng2021deepmulticap,zhang2021lightweight} or RGB-D camera arrays~\cite{yu2018doublefusion,su2020robustfusion}.
To reduce capture efforts, recent methods leverage deep neural networks to directly reconstruct 3d humans from even single images~\cite{natsume2019siclope,saito2019pifu,kolotouros2019learning,deng2019accurate}.
These methods often estimate output coefficients of parametric models of 3D human shape and poses~\cite{SMPL:2015}. The parametric model of 3D humans is often constructed from a large database of scanned shapes of different humans in a variety of poses and the rigged with a pre-defined skeleton to animate the human mesh.

\noindent{\textbf{Neural 3D Representations.}}
Recently, neural representation of 3D scenes has attracted considerable attention in the literature~\cite{mildenhall2020nerf,barron2021mip,sitzmann2020implicit,park2019deepsdf,deng2021deformed,chen2021snarf}. These methods exploit a neural network (usually multi-layer perceptrons) to represent implicit fields such as signed distance functions for surface or volumetric radiance fields, thus inherently encoding 3D information in a view-consistent manner.
Among those neural representations, NeRF~\cite{mildenhall2020nerf} (and its variants) has surpassed previous state-of-the-art methods on novel view synthesis tasks for static objects.
Some works also extended NeRF to handle general space-time dynamic scenes~\cite{pumarola2021d,xian2021space,park2021nerfies}.
Our method targets extending NeRF to model dynamic representation of 3D humans with the help of parametric 3D body mesh models.

\noindent{\textbf{Rigging NeRF.}}
A prevalent approach for representing dynamic humans with NeRF is to rig NeRF with articulated models.
Common articulation choices are 3D pose skeletons~\cite{noguchi2021neural,su2021nerf} and parametric 3D mesh models~\cite{guo2021ad,peng2021animatable,liu2021neural,peng2021neural}.
Our method utilizes a parametric 3D mesh model~\cite{bogo2016keep} for articulation.
While we are similar to previous and concurrent works~\cite{peng2021animatable,noguchi2021neural,liu2021neural,su2021nerf} by sharing the same goal of modeling dynamic human body with articulated NeRF representation, our method differs them in two aspects.
First, we attempts to simplify the input to monocular video input as opposed to multi-view video inputs~\cite{peng2021animatable,liu2021neural} and relax the dependence on accurate 3D geometry input~\cite{noguchi2021neural} a priori.
Second, we propose a new embedding method for querying articulated dynamic NeRF with locality and distance-preserving constraints.
Noguchi et al.~\cite{noguchi2021neural} proposed to learn a most relevant articulated part for any given query point.
The concurrent work of Su et al.~\cite{su2021nerf} propose a similar framework with joint-optimization of NeRF and human pose, using the skeleton as the human shape representation and directly relates the input query to all articulated skeleton joints. We focus on improving the generalization ability for NeRF-based animatible 3D human reconstruction with novel embedding designs. Our method preserves locality via nearest-neighbor projection, and encourages locality distance-preserving to avoid collapse of deformation in the whole volume.

\section{Method}
Given a monocular video sequence $\{\mathbf{I}_i\}_{i=1}^K$ as input, we aim to construct a neural human representation that encodes both appearance and geometry knowledge and can be rendered under an arbitrary pose $\theta$.
In particular, we model our representation with a neural radiance field (NeRF). 
Our NeRF is dynamically controlled by an underlying parametric mesh model (Sec.~\ref{sec:method:mesh_guided_nerf}). Given an observation-space pose, the mesh surface is deformed from its rest pose correspondingly. We design a novel query embedding (Sec.~\ref{sec:method:query_embedding}) for the input which encodes both information of surface deformation and addition constraints.
Based on the proposed mesh-guided NeRF, we propose an analysis-by-synthesis method to jointly estimate pre-frame 3D mesh from the input video and train NeRF (Sec~\ref{sec:method:joint_training}), using off-the-shelf tools for mesh initialization.

\subsection{Mesh-guided NeRF}
\label{sec:method:mesh_guided_nerf}
In NeRF, the rendered color $\bar{\mathbf{C}}(u,v)$ at image pixel $(u,v)$ is generated by querying and blending the radiance along the corresponding camera ray according to the volume density value: \begin{equation}
    {\bar{\mathbf{C}}}(u,v) = \sum_{i=1}^N{T_i(1-\exp(-\sigma_i\delta_i))\mathbf{c}_i},
    \label{equ:nerf_r}
\end{equation}
where
\begin{equation}
    T_i = \exp(-\sum_{j=1}^{i-1}{(-\sigma_j\delta_j}))),
    \label{equ:nerf_t}
\end{equation}
and
\begin{equation}
    (\mathbf{c}_i, \sigma_i) = F(\mathbf{x}_i).
    \label{equ:nerf_query}
\end{equation}
$\mathbf{c}_i \in \mathcal{R}^3$ and $\sigma_i$ are the color and volume density of the $i$-th sampled point $\mathbf{x}_i$ along the ray direction. $F(\mathbf{x})$ is usually parameterized with an MLP network.

We extend NeRF to handle the dynamic, articulated human body with a mesh-based parametric 3D model SMPL~\cite{SMPL:2015}. An SMPL model $S(\theta, \beta)$ takes a human 3D pose $\theta$ of skeleton joint rotations as well as a low-dimensional feature vector of human shape as input and outputs a 3D mesh. As we mainly focus on synthesizing humans under different poses, we omit the shape $\beta$ afterwards.

Formally, given a pose input $\theta$, the radiance color $\mathbf{c}(\mathbf{x})$ and volume density of our mesh-guided NeRF at point $x$ is computed as follows:
\begin{equation}
    (\mathbf{c}(\mathbf{x}), \sigma(\mathbf{x})) = F_{\Phi}(q(\mathbf{x}, S(\theta))),
    \label{equ:mesh_guided_nerf}
\end{equation}
where the query embedding $q$ is the most important part as it directly relates the output of NeRF with the underlay deformable mesh, as we will discuss next.

\subsection{Query embedding for NeRF}
\label{sec:method:query_embedding}
\begin{figure}[t]
    \centering
    \includegraphics[width=0.7\textwidth]{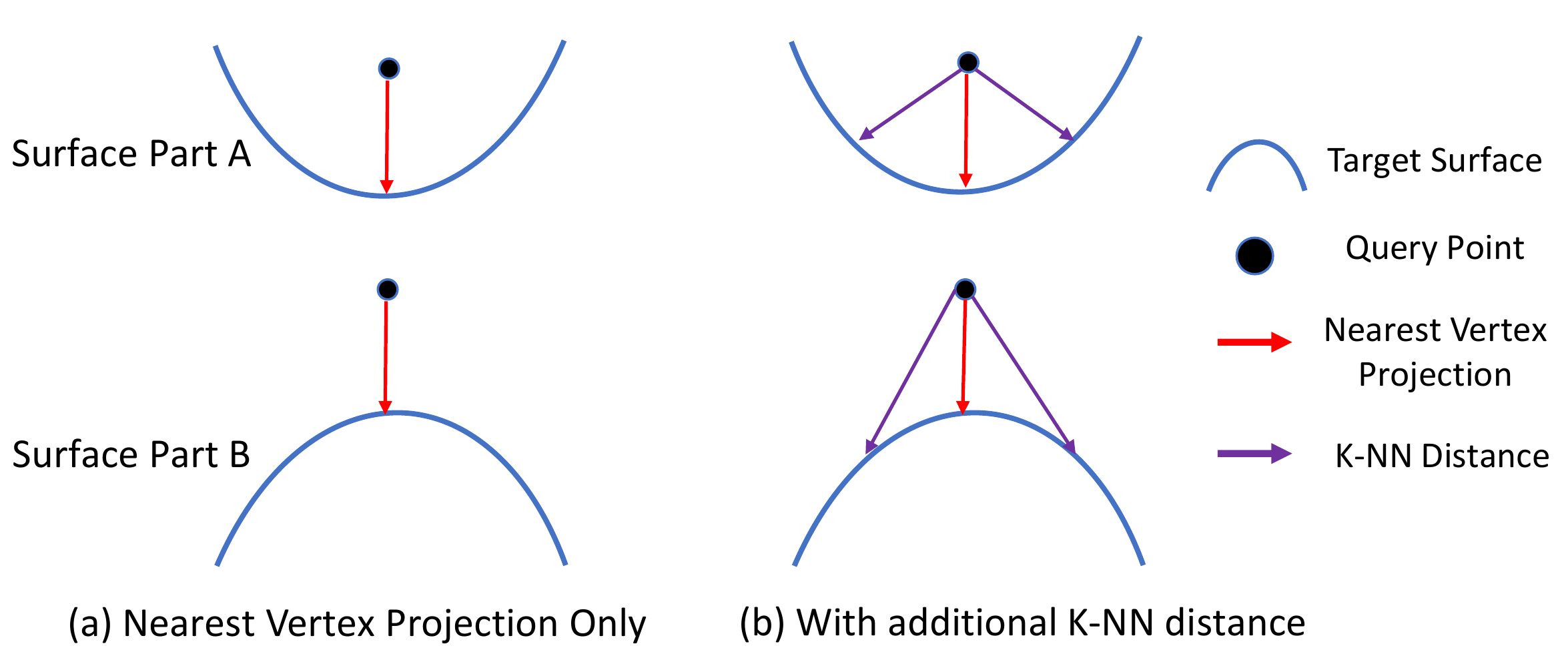}
    \caption{An illustration of our distance-preserved query embedding. (a) Naively embedding the query with nearest neighbor vertex on mesh (the red line), leads to indistinguishable embedding of different surface deformation patterns. (b) With additional geometric K-NN distance information (purple lines), different deformation patterns are clearly separated.}
    \label{fig:query_embedding}
\end{figure}

The input of NeRF for querying radiance value at point $\mathbf{x}$ is given by its 3D location $(x,y,z)$ and 2D viewing direction $\theta, \phi$ in the world space.
A natural extension of input querying for the dynamic scene is to define a deformation field that transforms observation-space points to rest space. Directly estimating a general deformation field together with the NeRF, as in~\cite{pumarola2021d,park2021nerfies,xian2021space}, is highly ill-posed and prone to local minima.
Inspired by~\cite{peng2021animatable,liu2021neural}, we leverage the deformable SMPL model as the human prior to guide our transformation for input queries.
The underlay SMPL model defines reasonable deformation fields on its surface; however, a radiance field from NeRF is defined on full 3D volume, and we still need to determine the deformation on unconstrained off-surface points. Naively projecting off-surface points to its nearest vertex point on the mesh is not optimal because the off-surface deformation will be collapsed, as illustrated in Fig.~\ref{fig:query_embedding}.

We address this issue from another perspective: instead of inputting an inverse-transformed point with an explicitly defined deformation field for querying NeRF, we construct a query embedding of the input point which encodes two types of information: (1) information that guides how the deformation field should roughly be (denoted as \textit{Deformation Guidance}), and (2) priors that prevent the deformation field from collapsed local minima (denoted as \textit{Deformation Priors}). 
The NeRF then implicitly learns a radiance field based on the input embedding.
Fig.~\ref{fig:pipeline} illustrates our design of query embedding.

\noindent{\textbf{Deformation Guidance.}}
Our deformation guidance is based on the underlay SMPL model.
For the SMPL model, the transformation relationship between a canonical-space surface point $\mathbf{v}$ and its observation space counterpart $\mathbf{v}'$ is given by the linear blend skinning (LBS) algorithm~\cite{lewis2000pose}:
\begin{equation}
    \mathbf{v}' = (\sum_{k=1}^{K}w(\mathbf{v})_kG_k)\mathbf{v},
    \label{equ:lbs}
\end{equation}
where $K$ is the number of human parts, $G_k \in SE(3)$ is the transformation matrix of the $k$-th part on the human skeleton, and $w(\mathbf{v})$ is the blend weight.

Intuitively, the guidance information from the SMPL model should neither be too global such that the network inadvertently relates the output to irrelevant articulated parts~\cite{zeng2020srnet,noguchi2021neural}, nor collapse to a single nearest neighboring point as the deformation field will remain unconstrained (Fig.~\ref{fig:query_embedding}). To this end, we build the deformation guidance part of the input query with the nearest projected vertex on the mesh as well as the k-nearest adjacent vertices of the projected vertex in rest space via inverse LBS as:
\begin{equation}
    q_{g}(\mathbf{x}) = (\mathbf{x}_{dir}, \mathbf{v}_0,\mathbf{v}_1,...,\mathbf{v}_k),
    \label{equ:query_guidance}
\end{equation}
where $\mathbf{v}_k = (\sum_{l=1}^{L}w(\mathbf{v}_k)_lG_l)^{-1}\mathbf{v}'_k$
and $\mathbf{v}'_k$ is the k-th nearest neighboring mesh point in the observation space. Note that, we additionally give the relative direction from query point $\mathbf{x}$ to its projected point $\mathbf{v}_0$:
\begin{equation}
    \mathbf{x}_{dir} = \mathcal{R}((\sum_{k=1}^{K}w(\mathbf{v})_kG_k)^{-1})\frac{\mathbf{v}_0 - \mathbf{x}}{\|\mathbf{v}_0 - \mathbf{x}\|_2}.
    \label{equ:query_dir}
\end{equation}
Here $\mathcal{R}$ denotes the rotational part of the transformation matrix.

\noindent{\textbf{Deformation Priors.}}
Our deformation guidance embedding $q_g$ itself is based on mesh surface only and insufficient to ensure a well-defined deformation field in the whole volume.
We therefore provide an additional part to the query embedding by equipping the input query with the Euclidean distances to its nearest points in the observation space:
\begin{equation}
    q_p(\mathbf{x}) = (d_0,d_1,...,d_k),
    \label{equ:query_prior}
\end{equation}
where $d_k = \|\mathbf{v}'_{k} - \mathbf{x}\|_2$.
Using the distance in the observation space is important as such information preserves the local difference under different poses and leads to a distance-preserved deformation field.

\noindent{\textbf{Appearance latent code.}}
To better capture the geometry and appearance detail which cannot be captured by surface mesh deformation, we additionally provide a learnable latent code $l_k$ defined on each mesh vertex:
\begin{equation}
    q_a(\mathbf{x}) = (\mathbf{l}_0,\mathbf{l}_1,...,\mathbf{l}_k).
    \label{equ:query_latent}
\end{equation}

The complete query embedding for the NeRF input is generated by feeding the concatenation vectors into a tiny 3-layer MLP network $\psi$:
\begin{equation}
    q(\mathbf{x}) = \psi(\mathbf{\gamma}(q_g(\mathbf{x})),\mathbf{\gamma}(q_p(\mathbf{x})), q_a(\mathbf{x})),
    \label{equ:query_full}
\end{equation}
where $\gamma$ denotes positional encoding as used in the original NeRF~\cite{mildenhall2020nerf}.

\subsection{Joint mesh estimation and NeRF training}
\label{sec:method:joint_training}
Training the mesh-guided NeRF from monocular video input requires paired data of input frames $\{\mathbf{I}_i\}$ and human mesh $\{\mathbf{M}_i\}$. State-of-the-art monocular video based human mesh reconstruction methods such as~\cite{kolotouros2019learning,kocabas2020vibe}~produce plausible results for human mesh estimation; however, they are still not accurate enough for training our NeRF as non-aligned mesh part to the image will give incorrect guidance and make the NeRF over-fitting to misaligned training poses.
Hence we opt to use the plausible mesh estimates provided by prior solutions as initialization, and jointly fine-tune the mesh with NeRF training. Practically, we choose to optimize the pose parameter $\theta^i$ for each training frame instead of per-vertex mesh offset, as it gives us enough capability to refine mesh-image misalignment without too much flexibility that overfits to local minima.

\noindent{\textbf{Training Objective.}} Our training is guided by the reconstruction error between the mesh-guided NeRF and the ground-truth frames over the whole video sequence as well as a regularization term penalizing too large deviation from the initial pose estimation $\theta_0$:
\begin{equation}
    \mathit{L} = \sum_{i}\sum_{u,v}{\mathit{L}_i(u,v)} + \lambda_p\sum_{i}\|\theta^i - \theta_0^i\|_2^2,
\end{equation}
and
\begin{equation}
    \mathit{L}_i(u,v) = \|\bar{\mathbf{C}}(u,v) - \mathbf{I}_i(u,v)\|_2^2, 
\end{equation}
where $\mathbf{I}_i(u,v)$ is the ground truth pixel value at $(u,v)$ from the $i$-th frame.
$\bar{\mathbf{C}}(u,v)$ is computed using Eq.~\ref{equ:nerf_r}, Eq.~\ref{equ:mesh_guided_nerf} and the proposed query embedding (Eq.~\ref{equ:query_full}).
\section{Experiments}
\label{sec:expr}
\subsection{Experimental Setup}
\noindent{\textbf{{Datasets}}}
We conduct experiments on different datasets as follows: 
\begin{itemize}
    \item People-Snapshot~\cite{alldieck2018video}: This dataset contains 24 subjects with monocular videos performing turning around. Among them, we choose female-1-casual, female-3-casual, male-1-sport and male-9-plaza for training.
    We remove the background of the video frames with ground truth silhouettes provided and resize the video to half-size (1080p$\rightarrow$540p). An initial mesh is provided in the data.
    \item DoubleFusion~\cite{yu2018doublefusion}: This dataset contains only a sequence of one man, where the actor performs more complex actions while turning around. Thus, we consider it not suitable for a quantitative benchmark and only use it to show qualitative comparisons on novel pose synthesis.
    The initial mesh is provided in the dataset using additional depth information.
    \item ZJU-MoCap~\cite{peng2021neural}: This dataset contains multi-view video sequences of 9 objects with 21 cameras. We choose a single view (subject 313 and subject 386 from camera 7) for training.
    \item Human3.6M~\cite{ionescu2013human3}: This dataset consists of a large number of 3D human poses and corresponding multi-view video sequences. We follow the same protocol as \cite{su2021nerf}, extracting every 64th frame of the videos. We train the model on the subject 9 and subject 11. For each video, we select camera 2 as the input view and employ
    SPIN~\cite{kolotouros2019learning} to estimate the initial mesh from video frames.
\end{itemize}
\noindent{\textbf{{Network Structure}}}
The network $\psi$ in the query embedding module is implemented with a 3-layer MLP with 128 channels.
The NeRF network $\phi$ is composed of an 8-layer MLP with 256 channels. 
In the position embedding module, we implement the tiny 3-layer MLP $\psi$ with 128 channels, and the NeRF module $\phi$ for rendering is composed of 8-layer MLP with 256 channels.
We apply a positional encoding of 10 frequencies to query embedding features except latent codes. 

\noindent{\textbf{{Training Details}}}
We utilize Adam optimizer~\cite{kingma2014adam} with learning rate of $1e-4$ for optimizing NeRF and latent code. The learning rate of body poses is set to $5e-4$ and $\lambda_p$ is set to 2.0.
For volumetric rendering we employ the coarse-to-fine ray sampling strategy of~\cite{mildenhall2020nerf}. We also constrain the sampled rays to be more focused on the human part in the image by sampling rays within the 1.2$\times$ padding bounding box of 2D keypoints with 70\% probability, and randomly sampled in the whole image with 30\% probability. Each sampled ray is discreted within $[z_{near} - 0.04,z_{far}+0.04]$, where $z_{near}$ and $z_{far}$ denote the nearest and farthest ray-point intersection with body mesh, respectively.
Our model is trained with a single Nvidia Tesla V100 32GB GPU, and the training approximately takes 60 hours to converge.
For datasets without background mask available, we either apply an off-the-shelf matting algorithm~\cite{lin2021robust} or jointly model the background during training. Please check the supplemental materials for details.

\noindent{\textbf{{Evaluation Metrics}}}
Peak-Signal-to-Noise Ratio (PSNR) and Structural Similarity Index Measure (SSIM) are used to evaluate image quality.

\begin{table}[ht]
    \centering
    \begin{subtable}[h]{0.50\textwidth}
        \centering
        \scalebox{0.8}{
        \begin{tabular}{c|c|c|c|c} 
        \toprule
        \multirow{2}{*}{}  & \multicolumn{2}{c|}{SSIM $\uparrow$ }  & \multicolumn{2}{c}{PSNR $\uparrow$ }   \\ 
        \cline{2-5}
          & Training & Novel & Training & Novel  \\ 
        \midrule
        w/o distance      & 0.935           & 0.906           & 29.18          & 27.51           \\
        canonical distance & 0.926          & 0.916          & 29.06          & 27.48           \\
        \midrule
        observation distance       & \textbf{0.980} & \textbf{0.973} & \textbf{35.87} & \textbf{34.75} \\
        \bottomrule
        \end{tabular}
        }
        \caption{Impact of distance embedding.}
        \label{table:distance}
    \end{subtable}
    \hfill
    \begin{subtable}[h]{0.45\textwidth}
        \centering
        \scalebox{0.8}{
        \begin{tabular}{c|c|c|c|c} 
        \toprule
        \multirow{2}{*}{}  & \multicolumn{2}{c|}{SSIM $\uparrow$ }  & \multicolumn{2}{c}{PSNR $\uparrow$ }   \\ 
        \cline{2-5}
          & Training & Novel & Training & Novel  \\ 
        \midrule
        w/o direction & 0.942        & 0.923      & 30.52         & 28.15       \\ 
        w/o inverse  & 0.959         & 0.921      & 32.14        & 28.12       \\ 
        \midrule
        full           & \textbf{0.980}        & \textbf{0.972}    & \textbf{35.87}         & \textbf{34.75}       \\
        \bottomrule
        \end{tabular}
        }
        \caption{Impact of direction embedding.}
        \label{table:direction}
     \end{subtable}
     \begin{subtable}[h]{0.9\textwidth}
        \centering
        \scalebox{0.8}{
        \begin{tabular}{c|c|c|c|c} 
        \toprule
        \multirow{2}{*}{}  & \multicolumn{2}{c|}{SSIM $\uparrow$ }  & \multicolumn{2}{c}{PSNR $\uparrow$ }   \\ 
        \cline{2-5}
          & Training & Novel & Training & Novel   \\ 
        \midrule
        Euclidean, 2 neighbors          & 0.950 & 0.935 & 32.35 & 30.69  \\
        Geodesic, 2-hop neighbors       & 0.962 & 0.954 & 32.74 & 31.97  \\
        Geodesic, only nearest neighbor & 0.961 & 0.938 & 31.87 & 30.07  \\
        \hline
        Geodesic 1-hop neighbors & \textbf{0.980} & \textbf{0.972} & \textbf{35.87} & \textbf{34.75} \\
        \bottomrule
        \end{tabular}
        }
        \caption{Impact of neighborhood selection.}
        \label{table:neighbor}
     \end{subtable}
     \caption{Ablation studies on (a) type of direction, (b) type of distances and (c) type of neighborhood selection for embedding construction.}
     \label{tab:embedding_ablation}
\end{table}
\subsection{Ablation Studies}
To validate the influence of our proposed query embedding, we conduct the ablation study on the People-Snapshot dataset and report quantitative results on both training and test (unseen) poses, from the following aspects:
\\

\noindent{\textbf{Neighborhood range}}: As we have discussed in Sec~\ref{sec:method:query_embedding}, the deformation guidance from the SMPL model should be neither too global nor too local. We verified this by conducting training with different ranges of mesh neighborhood. The results are shown in Tab.~\ref{table:neighbor}. Either increasing range (\textit{2-hop neighbors}) or \textit{only nearest neighbor} projected point leads to degraded performance, both for training and novel poses.
We also test a variant of our method by sampling K-NN point based on Euclidean distance (\textit{spatial K-NN}) instead of geodesic distance. The results are also degraded as it fails to aware human part connectivity (\eg, two adjacent points in Euclidean space might belong to distinct human parts).

\noindent{\textbf{Distance prior}}: We validate the importance of distance information in Tab.~\ref{table:distance}. We remove the distance feature in the \textit{w/o distance} model, and substitute rest-pose distance for observation-pose distance in the  \textit{canonical distance} model.
Obviously, without distance information, the results are significantly degraded and the difference between training and novel poses is increased.

\noindent{\textbf{Relative direction}}: The impact of relative direction embedding is demonstrated in Tab.~\ref{table:direction}, where \textit{w/o direction} denotes embedding without direction, and \textit{w/o inverse} denotes embedding direction in observation space. It is worth noting that the \textit{w/o inverse} greatly reduces the generalization on novel poses.

\noindent{\textbf{Pose refinement}}:
Our joint pose refinement with NeRF training is crucial when the initial mesh is not accurate enough.
To validate this, we conduct experiments on both Human3.6M and People-Snapshot dataset. The People-Snapshot dataset has provided an initial mesh that is rather reasonable; yet, we still observe minor artifacts without pose refinement and our joint training further improves the result, both quantitatively (Tab.~\ref{table:refinement}) and qualitatively (Fig.~\ref{fig:refinement}).
\begin{figure}[!ht]
    \centering
    \includegraphics[width=0.65\textwidth]{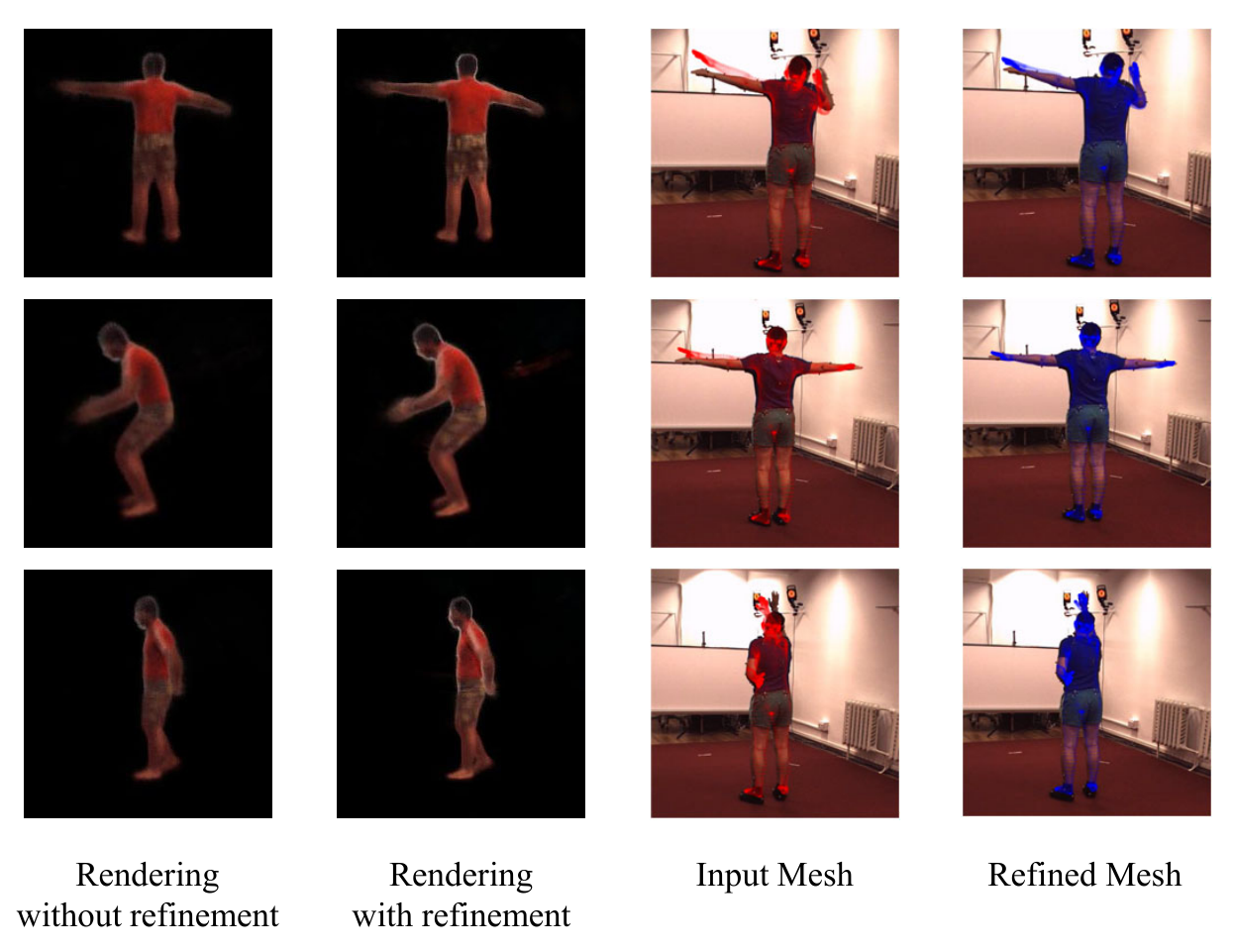}
    \caption{Qualitative comparison between original and optimized mesh. The final result corrects the initial human mesh, e.g., the alignment error on the arms.}
    \label{fig:refine_mesh}
\end{figure}
\begin{table}[htb]
   \begin{minipage}[b]{0.38\textwidth}
        \centering
        \scalebox{0.70}{
        \begin{tabular}{c|c|c|c} 
        \toprule
                            Dataset  & Method                & SSIM $\uparrow$    & PSNR $\uparrow$  \\ 
        \midrule
        \multirow{2}{*}{Human3.6M}       & w/ refinement  & \textbf{0.978} & \textbf{31.51}  \\
                                         & w/o refinement & 0.951          & 29.04          \\ 
        \midrule
        \multirow{2}{*}{People-Snapshot} & w/ refinement  & \textbf{0.972} & \textbf{34.75}  \\
                                         & w/o refinement & 0.969 & 32.99  \\
        \bottomrule
        \end{tabular}
        }
        \caption{The effect of using joint pose refinement.}
        \label{table:refinement}
   \end{minipage}\hfill
   \begin{minipage}[b]{0.57\textwidth}
        \centering
        \scalebox{0.70}{
        \begin{tabular}{c|c|c|c|c|c|c|c} 
        \toprule
             & \multicolumn{2}{c|}{ZJU-Mocap} & \multicolumn{3}{c}{Human3.6M} & \multicolumn{2}{c}{People-Snapshot}  \\
        \cline{2-8}
             & AniNeRF & Ours                 & AniNeRF & A-NeRF & Ours & AniNeRF & Ours                 \\
        \midrule
        SSIM $\uparrow$ & 0.758 & \textbf{0.768} & 0.865  & \textbf{0.928} & 0.912 & 0.948  &  \textbf{0.973}               \\
        PSNR $\uparrow$ & 23.75   & \textbf{25.01}               & 23.44 & \textbf{27.45} & 27.11 & 29.11   & \textbf{34.75}              \\
        \bottomrule
        \end{tabular}
        }
        \caption{Quantitative comparison with AniNeRF and A-NeRF.}
        \label{table:comparison}
   \end{minipage}
\end{table}

\subsection{Comparsions}
\label{sec:expr:results}
As there is very few (formally peer-reviewed and published) NeRF-based work that shares the same succinct \textbf{monocular} inputs with \textbf{mesh-based} geometry proxy as ours, we compare with the following methods:
\\

\noindent{\textbf{AniNeRF (ICCV 2021)}}
AniNeRF~\cite{peng2021animatable} is NeRF-based method for dynamic human modeling. AniNeRF also uses mesh as geometry guidance but requires more strict input requirements of multi-view video input. It produces high quality results with typically 3 to 4 synchronized views.
For a fair comparison, we follow the same single view setting and training data to re-train AniNeRF, and report the comparison results in Fig.~\ref{fig:zju_cmp}.
We emphasize that this experiment setup with monocular input is \textbf{not} for producing best-quality results, but to demonstrate the challenge of monocular video scenario as well as the benefit of our proposed method.
Compared with AniNeRF, our method generates complete skin and cloth, whereas AniNeRF is unable to model the whole body with limited view.
The quantitative result reported in Tab.~\ref{table:comparison} also shows our method outperforms AniNeRF under the same settings.
We also refer to the supplemental material for a comparsion to NeuralBody~\cite{peng2021neural}, the precursor method of AniNeRF.

\noindent{\textbf{A-NeRF (NeurIPS 2021)}}
A-NeRF~\cite{su2021nerf} is a recent work for modeling 3D human with NeRF using monocular video input. A-NeRF exploits joint optimization of NeRF with human skeletons. An apple-to-apple comparsion with A-NeRF is hard as it differs from our method in many implementation aspects which affects the result quality, e.g., from the underly parametric body representation (skeleton-based v.s. mesh-based) to the backbone capacities. Nevertheless, our result on the Human3.6M dataset is quantitatively comparable with A-NeRF (Tab.~\ref{table:comparison}).

\noindent{\textbf{Non-NeRF methods}
Regarding non-NeRF methods, we also compare our method with a SMPL-model based method, VideoAvatar~\cite{alldieck2018video}.
The qualitative results are shown in fig~\ref{fig:cmp_videoavatar}.
Given the same monocular video as input, the NeRF-based method generates results with more natural and realistic color effects.
\begin{figure}[!htb]
   \begin{minipage}[b]{0.64\textwidth}
     \centering
    \includegraphics[width=\textwidth]{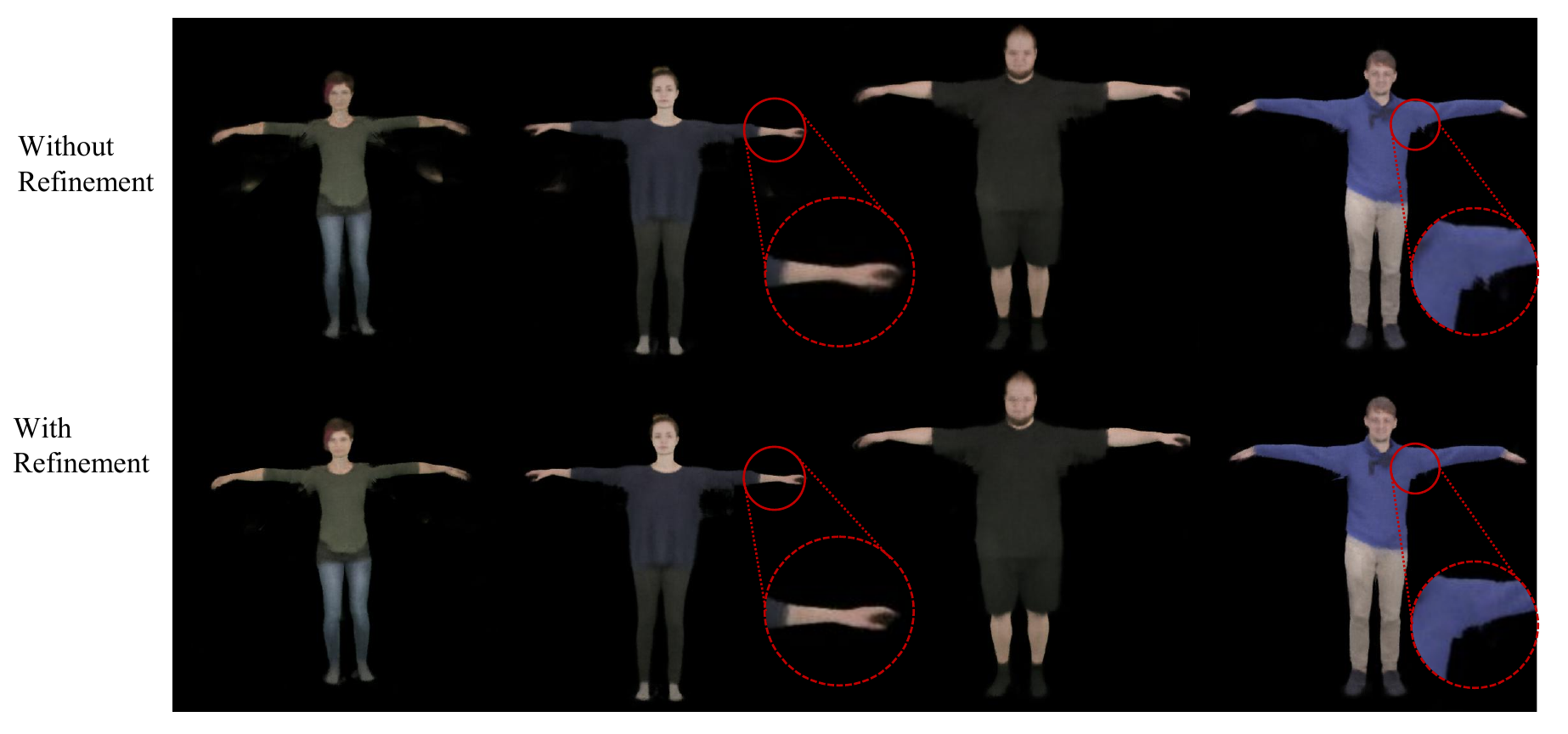}
     \caption{The effect of pose refinement on 
     People-Snapshot dataset. Jointly refinement contributes to clearer geometry and eliminates outliers. The improvement brought by refinement is enlarged in red.}
     \label{fig:refinement}
   \end{minipage}\hfill
   \begin{minipage}[b]{0.32\textwidth}
     \centering
     \includegraphics[width=\textwidth]{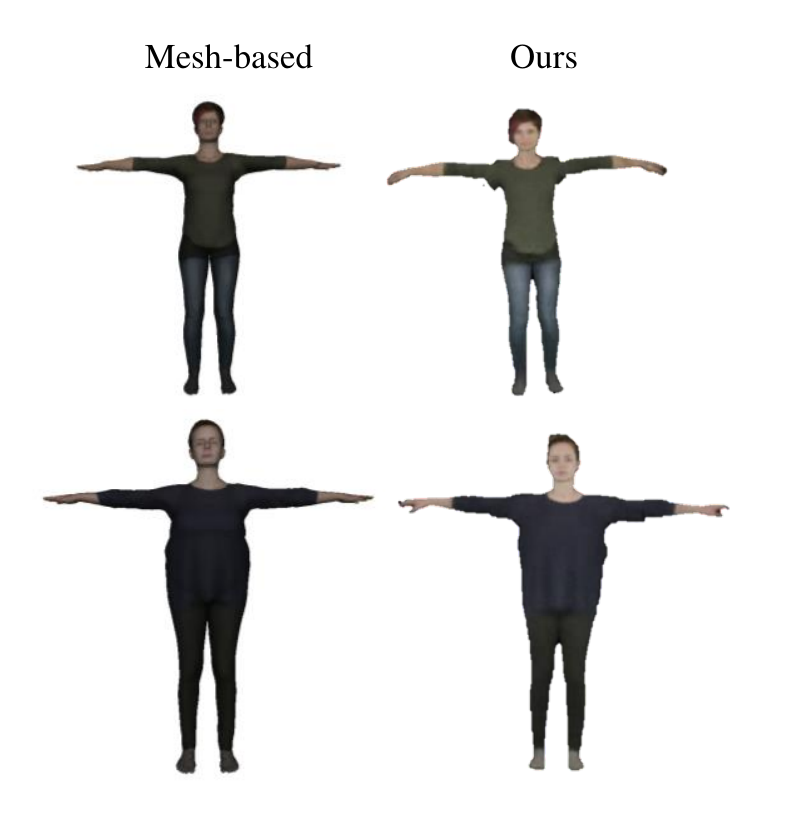}
     \caption{A qualitative comparison with mesh-based method VideoAvatar.}
     \label{fig:cmp_videoavatar}
   \end{minipage}
\end{figure}

\begin{figure*}[!ht]
    \centering
    \includegraphics[width=0.85\textwidth]{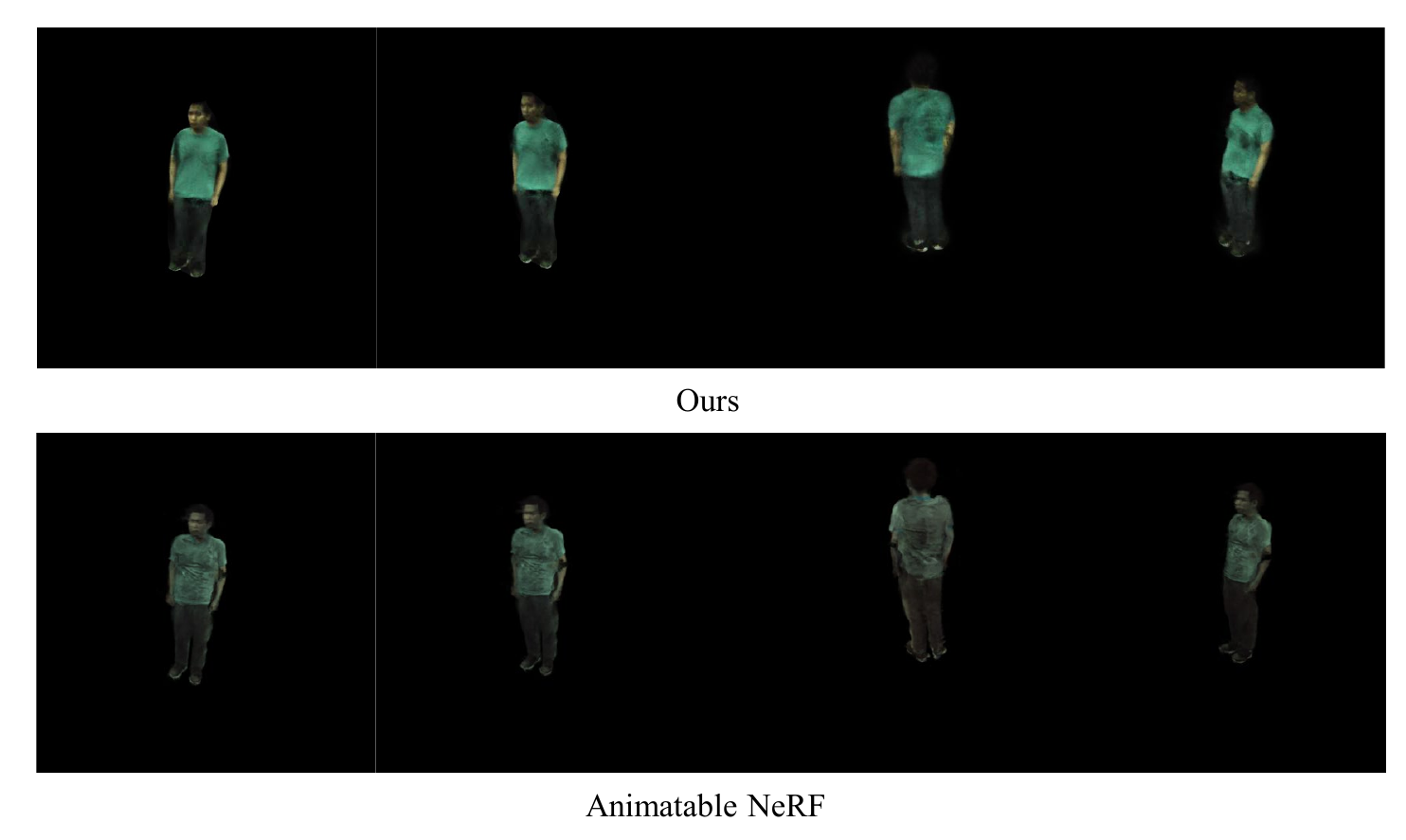}
    \caption{Qualitative comparison of Ani-NeRF~\cite{peng2021animatable} and ours under novel view. Both methods are trained with single view sequence.}
    \label{fig:zju_cmp}
\end{figure*}

\subsection{Applications}
\begin{figure}[ht]
    \centering
    \begin{subfigure}[b]{0.5\textwidth}
        \centering
        \includegraphics[width=\textwidth]{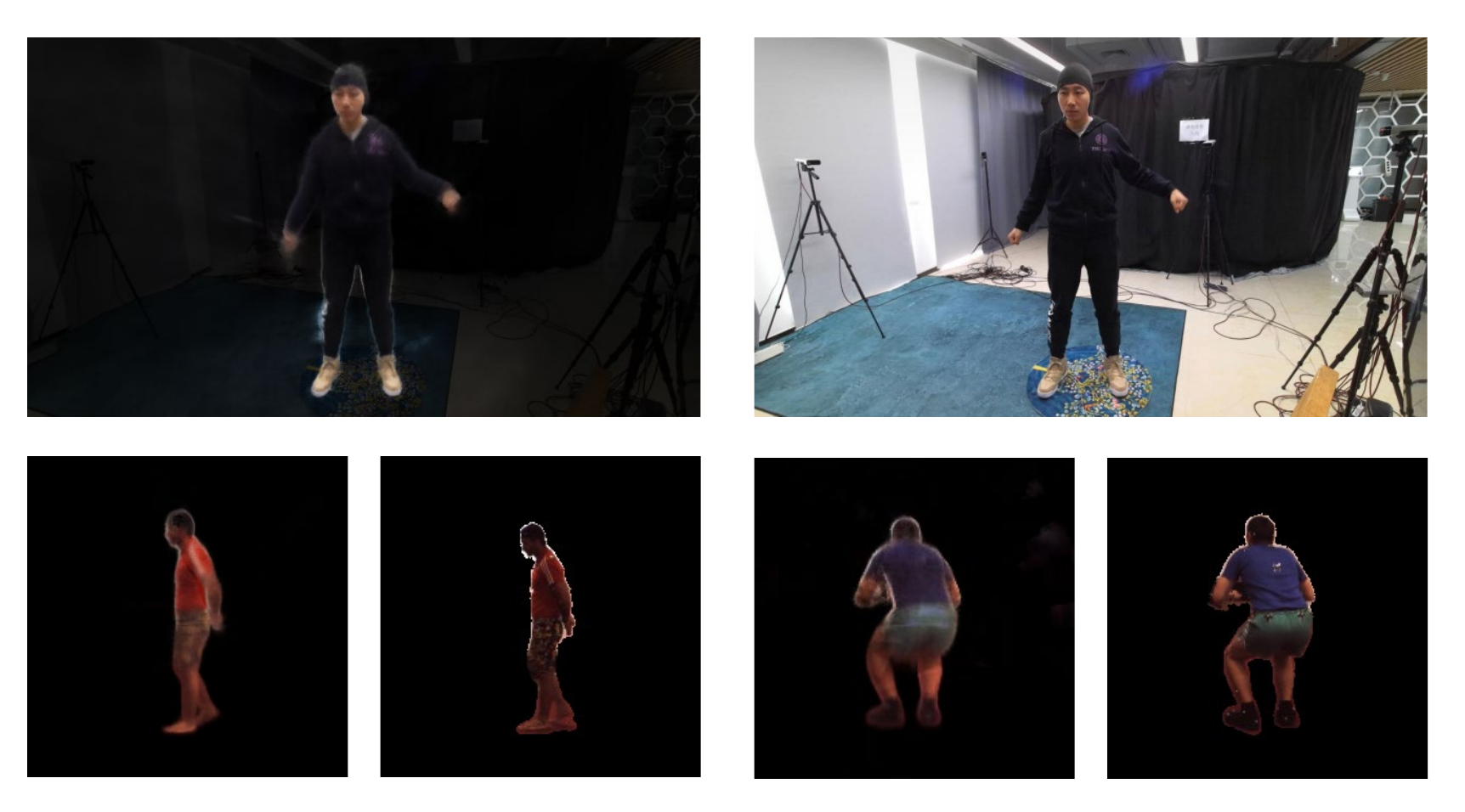}
        \caption{Qualitative results}
        \label{fig:novel_pose}        
    \end{subfigure}
    \hfill
    \begin{subtable}[b]{0.45\textwidth}
        \centering
         \scalebox{0.9}{
            \begin{tabular}{c|c|c|c} 
            \toprule
            \multirow{2}{*}{Metric} & \multicolumn{2}{c|}{Human3.6M\cite{ionescu2013human3}}        & People-Snapshot\cite{alldieck2018video}            \\
                            & \multicolumn{1}{c|}{S9} & \multicolumn{1}{c|}{S11}                      & Female-1                    \\
            \midrule
            SSIM $\uparrow$           &  0.978  &    0.972                & 0.973  \\
            PSNR $\uparrow$           &  31.51 &     29.70                   & 34.75   \\
            \bottomrule
            \end{tabular}
        }
        \caption{Quantitative results}
        \label{table:novel_pose}
     \end{subtable}
     \caption{Qualitative and quantitative results of \textbf{novel} pose synthesis on multiple datasets.
     (a) Top row: novel pose rendering (left) and ground truth (right) on DoubleFusion. Bottom row: rendering (odd column) and ground truth (even column) on Human3.6M. (b) Quantitative results of novel pose synthesis on Human3.6M and People-Snapshot dataset.}
\end{figure}

\begin{figure}[!ht]
    \centering
    \includegraphics[width=0.6\textwidth]{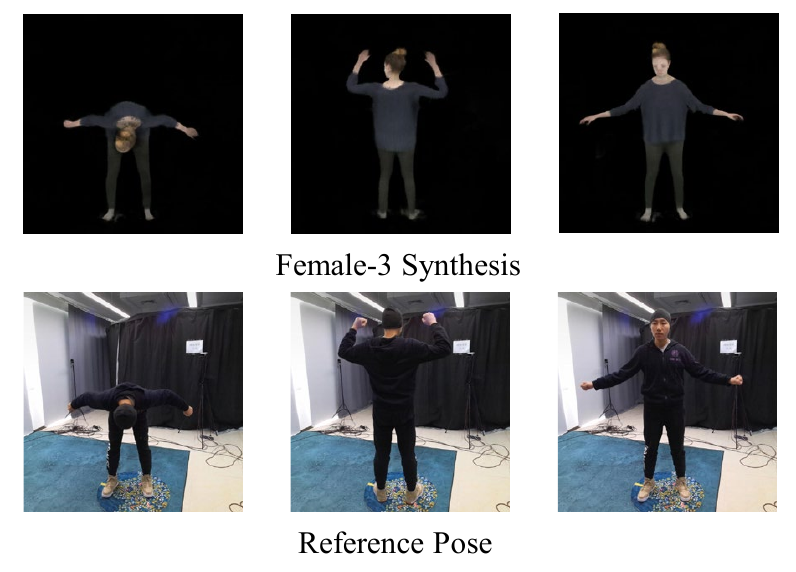}
    \caption{Human animation driven by Doublefusion poses. The synthetic human is trained on the People-Snapshot dataset.}
    \label{fig:novel_cross}
\end{figure}
\noindent{\textbf{Novel Pose Synthesis}}
Our trained representation enables character animation from novel unseen poses.
We evaluate our generalization ability by comparing testing data and our rendering driven by the same set of unseen poses on the People-Snapshot and DoubleFusion dataset.
The qualitative result is depicted in Fig.~\ref{fig:novel_pose}.
Our model successfully disentangles background and foreground pixels and veritably reconstructs the human body in the Doublefusion dataset (First row). As for side and back view, our model still generates images of high quality as shown in the Human3.6M dataset (Second row).
We also provide quantitative results in Tab.~\ref{table:novel_pose} on the People-Snapshot and Human3.6m datasets.

\noindent{\textbf{Pose Retargeting}}
The generalization ability of our model is further evaluated by pose retargeting experiments.
The results are shown in Fig.~\ref{fig:novel_cross}, where the driven poses derive from the Doublefusion dataset and training body comes from the People-snapshot dataset. We observe that our model generates realistic human bodys with various poses, which demonstrates the generalization of the proposed methods. We refer to the supplemental material for more novel pose synthesis results, including animation videos.

\section{Conclusion}
\label{sec:conclusion}
We presented a new method for building animatable neural 3D human representations from only monocular video inputs.
Our representation is based on dynamic Neural Radiance Field guided by parametric 3D human meshes. We designed a novel input query embedding of the mesh-guided NeRF.
We train the representation by we first initialize per-frame 3D meshes using off-the-shelf tools and then joint optimizing the 3D mesh and dynamic NeRF.
The learned neural representation can generalize well to unseen views and poses.
\\

\noindent{\textbf{Limitations}}
Our method is not without limitations.
The input embedding of our querying is related to a local region on the mesh surface with a restricted reception field; thus the joint optimization might fail if the initial pose has deviated too much from the ground truth.
Due to resolution constraint and the expressiveness of the mesh model we used, our method is still straggling at recovering high-resolution details such as human faces.
\\

\noindent{\textbf{Future work}}
For future works, we plan to explore different kinds of deformation priors and their effects on rigging dynamic NeRF, improving our performance with sharp details, and extending to general, non-articulated dynamic objects.
\\

\noindent{\textbf{Acknowledgements}} We would like to thank all the reviewers for their constructive feedback.

\clearpage
{\small
\bibliographystyle{splncs04}
\bibliography{egbib}
}

\end{document}